# An Effective and Efficient Approach for Clusterability Evaluation

Margareta Ackerman, Andreas Adolfsson, and Naomi Brownstein

Florida State University

Abstract

Clustering is an essential data mining tool that aims to discover inherent cluster structure in data. As such, the study of clusterability, which evaluates whether data possesses such structure, is an integral part of cluster analysis. Yet, despite their central role in the theory and application of clustering, current notions of clusterability fall short in two crucial aspects that render them impractical; most are computationally infeasible and others fail to classify the structure of real datasets.

In this paper, we propose a novel approach to clusterability evaluation that is both computationally efficient and successfully captures the structure of real data. Our method applies multimodality tests to the (one-dimensional) set of pairwise distances based on the original, potentially high-dimensional data. We present extensive analyses of our approach for both the Dip and Silverman multimodality tests on real data as well as 17,000 simulations, demonstrating the success of our approach as the first practical notion of clusterability.

## 1 Introduction

Clustering is a ubiquitous data analysis tool, applied in virtually all disciplines. The goal of clustering is to uncover meaningful cluster structure in data. As such, the application of this data mining tool relies on the presence of inherent structure, making notions of clusterability, which aim to quantify the degree of cluster structure in data, integral to cluster analysis.

Evaluation of clusterability plays a central role in both clustering theory and practice. Clusterability analysis should precede the application of clustering algorithms, because the success of any clustering algorithm depends on the presence of cluster structure.

To see how clusterability fits within the clustering process, consider the clustering pipeline depicted in Figure 1.[1] The process begins with data preprocessing, often involving feature selection or extraction. Next, clusterability analysis determines whether the data possesses sufficient cluster structure. If it is discovered that data does not possess sufficient cluster structure to be meaningfully partitioned, then clustering may not be suitable for the given data, or the data may be reprocessed.

---

[1] A similar pipeline is presented in the famous survey of clustering algorithms by Xu and Wunsch [35], sans the third step. Figure 1 shows how clusterability fits within the clustering process.

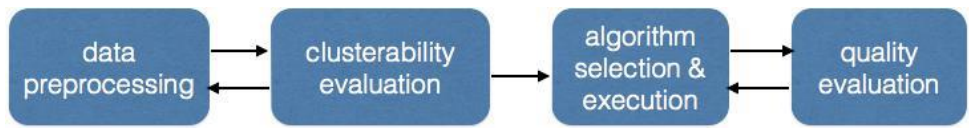

Figure 1: Clustering pipeline. This figure shows the feedback pathways in cluster analysis and role of clusterability in this process.

On the other hand, if the data is found to be clusterable, a suitable algorithm may be selected or developed.[2] After the algorithm is executed, the solution is validated by applying clustering quality measures [1, 30], which may result in the selection of an alternate algorithm.

In the theoretical analysis of clustering, data clusterability is an essential consideration, as many desirable characteristics of clustering algorithms arise only when the data is sufficiently clusterable. Since the results of clustering methods are often inconsequential when data fails to possess inherent cluster structure, it suffices to study how algorithms behave on clusterable sets. Clusterability notions have been used to show that, when data possesses inherent cluster structure, various algorithms are computationally efficient [12, 32], produce the desired output [11, 7], and possess other desirable characteristics, such as robustness to small adversarial sets [6].

While prior notions of clusterability help gain insight into the behavior of clustering techniques, they fall short from a practical standpoint. Perhaps the most significant hindrance to the application of prior notions is that most are NP-hard to compute [2]. The few exceptions that can be computed in polynomial time fail to capture the structure of real datasets, as they are designed to identify exceptionally well-structured data that rarely occur in practice, as detailed in Section 1.1.

Computational infeasibility and failure to identify the structure of real datasets inhibit the practical application of prior notions of clusterability. Further, inability to apply these notions in practice ultimately limits the implications of theoretical results relying on these notions, making it difficult to gauge whether or when these notions capture how real data is structured. Thus, both clustering theory and practice stand to benefit from a practical notion of clusterability.

Lastly, many previous notions of clusterability are based on specific algorithms or objective functions [32, 11, 12, 8, 2, 9]. This effectively inverts the clustering pipeline, requiring that we choose an algorithm before we determine whether data possesses sufficient structure to be meaningfully clustered. Further, since different clustering algorithms identify distinct types of cluster structure [7, 6], relying on a specific algorithm restricts a notion of clusterability to identifying structure that the underlying algorithm can capture.

In this paper, we propose the first practical notion of clusterability. Our method of clusterability evaluation describes the structure of real sets, runs is low polynomial time, and is independent of any algorithm or objective function. In an effort to develop this notion, we make an important paradigm shift. Until now, notions of clusterability were typically developed in the theoretical realm [32, 11, 12], with the aim of identifying when algorithms exhibit interesting behavior. Ours is the first notion developed through extensive data analysis.

By developing a clusterability measure that can be computed quickly and captures the structure of real data, we allow users to start utilizing clusterability for what is arguably its primary purpose: to discover whether data possesses sufficient cluster structure to meaningfully apply clustering techniques. Further, theoretical analysis relying on our notion of clusterability benefits from its ability to capture the structure of real sets.

The key to our approach lies in the discovery that a dataset's histogram of dissimilarities reveals a lot of information about its inherent cluster structure. Namely, the presence of multiple modes in the (one-dimensional) set of pairwise distances indicates that the original (possibly high-dimensional) data possesses inherent cluster structure. We show how tests of multimodality, namely the Dip and Silverman tests, can be applied on the set of pairwise dissimilarities to determine whether the underlying data is clusterable. Notably, our approach is computationally efficient, being quadratic in the input size, which is a radical improvement over standard NP-hard approaches.

The remainder of the paper is organized as follows. We begin with an overview of previous work on clusterability and an introduction to tests of multimodality. We then present a detailed description of our approach and several illustrative examples. Next, we describe our extensive simulations and findings on real data. We conclude with a discussion and future work.

---

[2]Note that multiple methods should be considered at this step, because different algorithms are apt at identifying different types of cluster structures [5, 7].

## 1.1 Previous work

Due to their central role in cluster analysis, many different notions of clusterability have been proposed. Most of these notions rely on a specific, often NP-hard, objective functions. For instance, one elegant notion defines data as clusterable when clusterings of near-optimal cost possess near-optimal structure [11]. Another similar notion considers the ratio of the optimal k-means solution over the optimal solution with k-1 clusters to attain a clusterability score [32]. Other notions that rely on optimal solutions for NP-hard objective functions were considered in [8, 2] and [9].

Another class of clusterability notions consists of those that can be efficiently computed, but are too strict for most practical applications. For example, Epter [20] defines data as clusterable when the minimum between-cluster separation exceeds the maximum in-cluster distance. Another elegant notion, by Balcan et al [12], defines data to be clusterable when each element is closer to all elements in its cluster than to all other data.

Both the computationally infeasible notions of clusterability and those that are too strict for practical applications helped enrich our understanding of clustering and played a key role in many recent findings [12, 3, 32, 6, 11, 4]. However, they leave open the challenge of developing a realistic notion for evaluating the degree of inherent cluster structure in data, which is the problem we tackle in the current work.

Detecting the number of clusters is a related problem, where the typical approach involves the analysis of clustered data, requiring that we select and execute an algorithm before determining whether multiple clusters are present [16]. This inverts the clustering pipeline, typically requiring numerous runs of both the algorithm and a validity index, but most importantly, this approach may fail to detect multiple clusters if an inappropriate algorithm is chosen, as different clustering algorithms are apt at detecting different cluster structures [5, 6, 7]. As such, selecting an inappropriate algorithm may occlude the presence of multiple clusters in the data, leading to invalid conclusions.

Another related measure that can be viewed as a method for evaluating clusterability is stability [15], which evaluates the consistency of an algorithm's behavior on different samples of the same data. Like typical approaches for detecting the number of clusters, whether data is found to be clusterable using stability depends on the chosen algorithm. By contrast, we seek to test whether data is clusterable independently of any choice of clustering algorithm. This allows one to determine whether clustering is warranted.

Despite its prominence as a central concept in clustering, very little work has been done studying clusterability as an independent construct. The first theoretical study of clusterability by Ackerman and Ben-David [2] compared previous notions, showing that they are pairwise inconsistent; for each pair of notions considered, there was data for which one of the notions evaluated as well-clusterable, while another indicated that the data is poorly clusterable. This highlights the importance of treating notions of clusterability with care, as notions which seem natural at first glance are often radically different from one another.

Ackerman and Ben-David [2] found that whenever data is clusterable using any of the studied notions, there is an algorithm that can be used to efficiently cluster the data. Daniely et al [19] recently studied this phenomenon, proposing the hypothesis, also known as the CDNM thesis, that "clustering is difficult only when it does not matter." Ben-David [14] more carefully considered this hypothesis, concluding that the CDNM thesis is far from being substantiated.

In an effort to explore the CDNM hypothesis, Ben-David [14] brought up the open problem of developing novel notions of clusterability, proposing four properties that these notions should satisfy. Among these four requirements are the following two which pertain to the clusterability evaluation component of the pipeline in Figure 1: (1) It should be reasonable to assume that most (or at least a significant proportion of) the inputs one may care to cluster in practice satisfy the clusterability notion and (2) There exists an efficient algorithm for testing clusterability; Namely, given an instance, the algorithm determines whether it satisfies the clusterability requirement or not. No previous methods satisfy both of these conditions, and there is no evidence that any satisfy (1). In this paper, we introduce the first approach to clusterability evaluation that satisfies these two requirements.

## 1.2 Multimodality tests

Determining whether a dataset has two or more modes is an integral component of our approach. A number of tests for multimodality have been developed. One popular test by Silverman [34] is based on the kernel density estimate. The general idea involves approximating the empirical distribution of the observed data with a set of Gaussian distributions of a specified bandwidth. If a sufficiently large bandwidth is required to produce a unimodal estimate, then the test concludes that the data distribution is multimodal. Another, the Dip test, compares the empirical distribution to a uniform density and rejects the assumption of unimodality if the data is sufficiently different from the closest possible uniform distribution [25]. Tests of multimodality have been applied to studies of carbon dixoide emissions, income levels, body size, cognitive processes, and archeological artifacts [26, 35, 23, 13, 18].

## 2 Method for clusterability evaluation: Our approach

Developing measures of clusterability has thus far been primarily considered in the theoretical context, often with the goal of demonstrating that specific algorithms possess desirable characteristics. We take a fresh perspective at the development of clusterability measures, proposing a new approach for studying and evaluating clusterability. This section focuses on an exposition of our approach, while the following two sections provide extensive simulations and results from real datasets.

The main insight in our approach is the observation that clusterability can be inferred from a one-dimensional view of the pairwise distances. That is, the lengths of the pairwise distances are sufficient for this analysis, without the need for considering how these distances are arranged to form the data.

To illustrate this approach, we begin with two extreme cases. First, consider highly clusterable data, consisting of two very well-separated, dense regions as depicted in Figure 2b. This dataset consists of two types of pairwise distances: very long ones, corresponding to between-cluster distances, and very short ones, corresponding to within-cluster distances. The histogram of pairwise distances depicted in the right column of Figure 2b reveals a sharp drop-off, clearly delineating the breaking point between in-cluster and between-cluster distances and resulting in two clear modes.

In general, when the underlying data is clusterable, it leads to a multimodal distribution of the pairwise distances. However, it is important to note that the modes do not represent distinct clusters. Instead, they tend to group together in-cluster and between-cluster distances.

Now, consider an example of unclusterable data, such as the random data generated using a single Gaussian distribution, depicted in Figure 2a. For this data, the histogram of pairwise distances is unimodal.

While the above examples consider clear cases of clusterable and unclusterable data, our findings include a large number of experiments that confirm this phenomenon: The degree of clusterability is captured in the histogram of its pairwise distances, allowing us to determine the degree of clusterability by counting the number of modes found in this histogram. In particular, multiple modes indicate that data is clusterable, while a single mode signals that data is not clusterable.

Figures 2c-f illustrate additional key examples, showing that our findings persist in the presence of noise (Figure 2d), varied cluster diameters (Figure 2e), different cluster sizes (Figure 2f), and even when the separation between clusters is small (Figure 2c). Analysis of clusterable data consistently uncovers a fall-off after the first mode, followed by one or more additional modes. In stark contrast, unclusterable sets consistently yield unimodal dissimilarity distributions. The next two sections analyze many additional datasets.

While the method of constructing a dendrogram of pairwise distances remains of independent interest, we propose applying multimodality tests that provide concrete recommendations and eliminate the need the need for creating dendrograms. To this end, we utilize statistical tests of multimodality on the set of pairwise distances, which are inherently one-dimensional. In particular, we use the Dip and Silverman tests. The following sections demonstrate the application of both tests on simulated and real data.

The Dip and Silverman tests each provide a p-value, which indicates the probability of seeing the given input or a more extremely multimodal input if the data is unimodal. If only a single mode is present, then the p-value should be large.

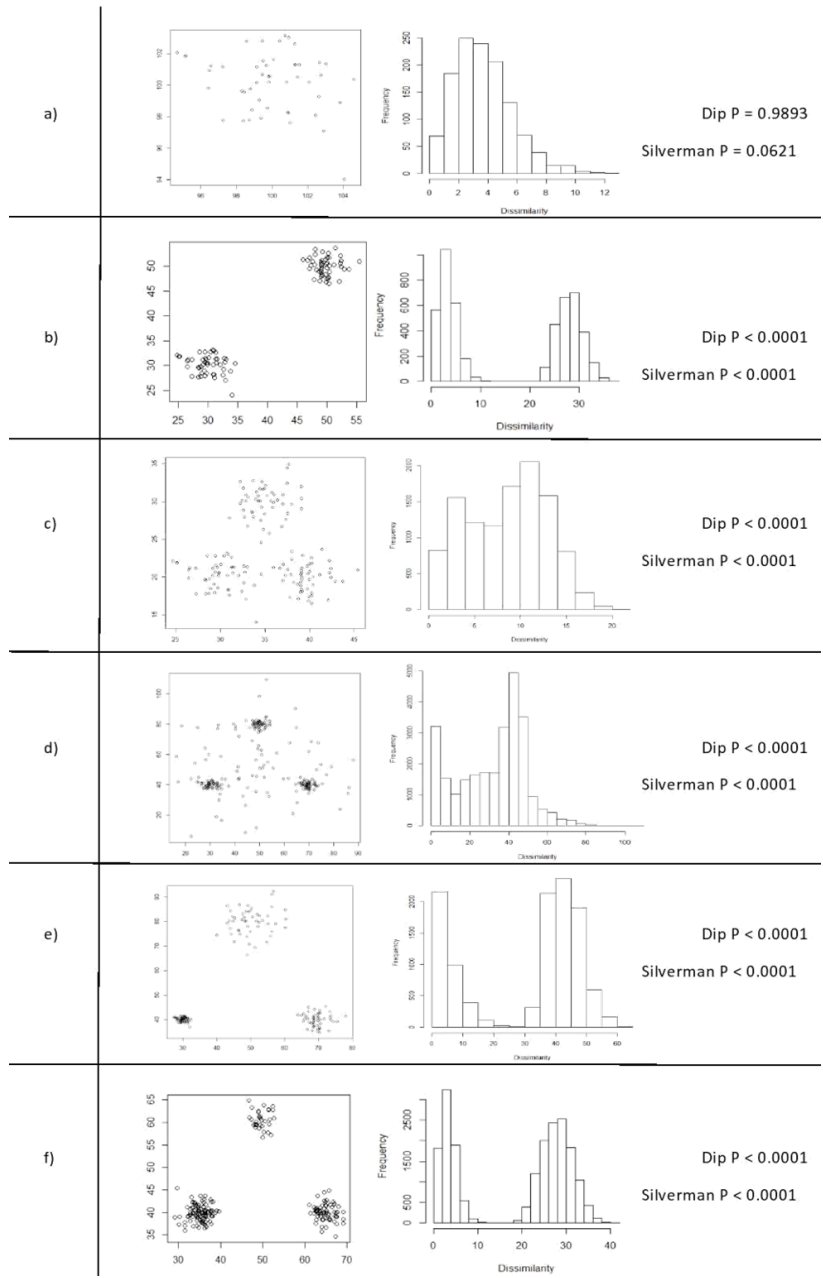

Figure 2: Illustrative examples demonstrating the foundation of our approach for clusterability evaluation. Figure (a) depicts unclusterable, random data generated using a single Gaussian distribution, along with the corresponding histogram of the underlying pairwise dissimilarities, which consists of a single mode. In contrast, Figure (b) shows data that possesses clear clustering structure, comprised of two dense, well-separated, clusters, with a corresponding bi-modal histogram of pairwise dissimilarities. Figures (c) through

(f) demonstrate the robustness of our approach in the presence of noise, when cluster diameters and sizes vary, and even when cluster separation is small. The p-values attained for the dip and Silverman multimodality tests are displayed next to each dataset, signaling that data is clusterable whenever $p < 0:05$ and indicating that data is unclusterable otherwise.

This indicates that the underlying data is not clusterable. On the other hand, small p-values make us question the original assumption of unimodality and instead conclude that multiple modes are present in the population. Consequently, this lets us conclude that the underlying data is clusterable. Note that both tests run in low polynomial time, where Silverman's test is quadratic, while the Dip test is linear in the number of pairwise distances [28].

The user should a priori choose a cut-o score, or significance level, to determine the point below which datasets would be determined to be clusterable. For example, the significance level of 0:05 is common. In this case, over the long term, we would expect 5% of unclusterable datasets to be (incorrectly) classified as clusterable. One could choose a smaller significance level at a cost of lower statistical power to correctly classify truly clusterable datasets as clusterable. By convention, we use the significance level of 0:05, where values below 0:05 allow us to conclude that the distribution of pairwise distances has multiple modes.

Figure 2 includes the Dip and Silverman p-values for each of the depicted datasets. Note that both multimodality tests correctly identify the data in Figure 2a as unclusterable, with p-values greater than 0:05, and the other datasets in this figure as clusterable, by having p < 0:05. The following two sections present extensive experimental results on both simulations and real datasets, confirming the validity of our approach.

## 3 Simulations

We now describe our extensive simulations for evaluating our approach to clusterability using both the dip and Silverman tests. The simulations consist of 17 types of datasets, where each type was generated with the same parameters 1000 times, for a total of 17,000 simulations. The code is found in the supplementary material. For example, row (c) of Table 1 describes the results for all simulations with 3 clusters with small separation in 2D, for which we have selected parameters that generate 3 bivariate Gaussian clusters each consisting of 50 points, with means at (30; 20), (40; 20), and (35; 30). One simulation of this type is displayed in Figure 2c. Note that many of our tests are in 2D to facilitate visual classification of the data as clusterable or unclusterable. For high dimensional data, we examine all 2D projections.

We generate 1000 different datasets using these parameters. For each dataset, we run both the Dip and Silverman tests. In Table 1, we record the percent of datasets on which each of the tests has a p-value less than 0:05 (that is, the percent of time that the tests rejected the null hypothesis of unimodality at the traditionally used 0.05 significance level[3]). As such, high values in Table 1 indicate high values of clusterability, while low values indicate poor clusterability.

Row (c) of Table 1 shows that for the 1000 datasets consisting of 3 close clusters in 2D, 100% of these datasets have p < 0:05 for Silverman's test, and 99.7% of them have p < 0:05 for the Dip test. The Dip and Silverman tests correctly identify data of this type as clusterable 100% and 99:7% of the time, respectively. We now discuss the data visualizations and results of our findings. For full details on the how each simulation type was generated, please see the appendix.

### 3.1 Visualizing the data

As there are no previous practical notions of clusterability, visualizing the data was helpful for assessing the utility of our new approach. As such, most datasets were generated in 2D. Our simulations also show that our approach works in high dimension.

Observe that because we generate each of the 17 types of data 1000 times, it is unrealistic to include all visualizations. However, we display one visualization for each of the 17 types, showing 2D projections for the high-dimensional data. While space constraints necessitate placing most of these visualizations in the appendix, 6 of the 17 types appear in Figure 2. In particular, rows (a) through (f) in Table 1 correspond to the same rows in Table 2.

---

[3]For unclusterable datasets, the proportion of rejections corresponds to type I error, or the rate of erroneously classifying unclusterable datasets as clusterable. For clusterable datasets, the proportion of rejections corresponds to the statistical power, or ability of the test to correctly classify clusterable sets as having cluster structure.

## 3.2 Summary of results

We examine the results summarized in Table 1, beginning with clusterable data that consists of multiple Gaussian clusters. Rows (b) through (l) of Table 1 summarize these results, with different numbers of clusters, ranging from 2 to 10, either 2, 3, or 10 independent dimensions, varying degrees of separation between the clusters, and different cluster sizes and diameters. On most clusterable datasets, both Dip and Silverman correctly identified that the data is clusterable 100% of the time, as represented with a score of 1 in the table. In rows (c), (d) and (l) we find scores of 0:979 to 1:000 for dip test, and 0:997 to 1:000 for the Silverman test. This means that on 3 out of 1000 datasets with "3 close clusters in 2D," the dip test incorrectly identified the data as unclusterable (and similarly for the other two cases). In summary, the power for our simulations ranged from 97.9% to 100%.

Consider Table 1(a) and (m)-(q), summarizing the results of our simulations for single cluster data. Table 1(a), (m) and (n) all consist of a single Gaussian distribution, in dimensions 2, 3, and 10 respectively. Both Dip and Silverman generally conclude that the given data is unclusterable with excellent accuracy, namely 100% for the Dip test and over 95% for Silverman.

When outliers are introduced to unclusterable data, we observe a disparity between Dip and Silverman. In particular, rows (o), (p), and (q) represent a single Gaussian cluster and a small number of outliers. Here we find that the dip test identifies such data as unclusterable, while Silverman classifies it as clusterable.

Where the dip test is robust to outliers, the Silverman test allows for small clusters. This goes back to the inherent ambiguity of clustering; for some applications, small clusters are acceptable, while for others, robustness to outliers is desired. In fact, the same phenomenon is observed with clustering algorithms, where some tend to identify small clusters, while others effectively view such data as outliers [6].

In sum, our simulations indicate that the use of multimodality tests on the set of pairwise distances is an effective and accurate method of classifying datasets by their level of clusterability. Both clusterable and unclusterable datasets were identified as such in nearly all the simulations. Different behavior in the presence of outliers allows users to select which test to use based on how they prefer to treat small clusters.

Lastly, we note that we would have liked to compare our new notion of clusterability with previous ones; however, in this respect there is a major difficulty. As discussed in Section 1.1, prior notions are impractical due to unreasonable computational complexity or unrealistic requirements for clusterable data. See Section 5 of the appendix for more detail.

## 4 Results on real datasets

In this section, we demonstrate the success of our method of clusterability evaluation by analyzing real datasets from the R datasets package, and compare the Dip and Silverman tests on these sets. We present clusterable and unclusterable sets and the corresponding p-values from applying the Dip and Silverman tests on the sets of their pairwise distances. To determine whether our approach was able to evaluate clusterability appropriately in these real datasets, we manually judge the clusterability of the data through two-dimensional projections of the original data.[4]

All of the datasets we present are freely available within R and were selected to ensure sufficient sample size and varied dimensionality. We provide examples of multi-dimensional data for applicability to real-world problems and 2D examples for ease of visualization.

We begin by considering the famous iris dataset [22], depicted in Figure 3, which is known to have three clusters represented by three species of the iris flowers. The dataset gives the measurements in centimeters for the sepal length and width and petal length and width, respectively, for 50 flowers from each of the 3 species. Running the Dip test on the set of pairwise distances of the iris data, we obtain a p-value less than 0.0001 for both the Dip and Silverman tests, indicating that both methods indeed classify the data as clusterable.

Next, we consider the faithful dataset [24, 10], which captures waiting time between eruptions and the duration of the eruption for the Old Faithful geyser in Yellowstone National Park. This famous data consists of 272 observations on 2 variables, numeric eruption time in minutes and numeric waiting time to next eruption.

---

[4] We include the scatter plots for two of the datasets in the main body of the paper. The visualizations can be attained using R with the command plot (dataset name). We also include all plots in the appendix.

|   | Data | Dip | Silv. |
|---|---|---|---|
| a. | 1 cluster 2D | 0.000 | 0.038 |
| b. | 2 separated clusters 2D | 1.000 | 1.000 |
| c. | 3 close clusters 2D | 0.997 | 1.000 |
| d. | 3 noisy clusters 2D | 1.000 | 0.997 |
| e. | 3 clusters, varied diameters 2D | 1.000 | 1.000 |
| f. | 3 clusters, varied density 2D | 1.000 | 1.000 |
| g. | 3 separated clusters 2D | 1.000 | 1.000 |
| h. | 3 separated clusters 3D | 1.000 | 1.000 |
| i. | 2 separated clusters 10D | 1.000 | 1.000 |
| j. | 4 separated clusters 10D | 1.000 | 1.000 |
| k. | 10 separated clusters 2D | 1.000 | 1.000 |
| l. | 10 close clusters 2D | 0.979 | 1.000 |
| m. | 1 cluster 3D | 0.000 | 0.042 |
| n. | 1 cluster 10D | 0.000 | 0.036 |
| o. | 1 cluster 2D with outlier | 0.000 | 0.985 |
| p. | 1 large cluster 2D with outlier | 0.000 | 0.965 |
| q. | 1 cluster 2D with 3 outliers | 0.082 | 0.976 |

Table 1: Proportion of datasets classified as clusterable over 1000 runs for each type, for a total of 17000 simulations. High scores indicate highly clusterable data, and low scores indicate lower cluster structure.

As depicted in Figure 20 in the appendix, this real, two-dimensional dataset possesses exceptionally clear cluster structure. Both the Dip and Silverman tests agree with this conclusion ($p < 0.0001$).

The swiss dataset [31] consists of 6 standardized fertility measures and socio-economic indicators represented as percentages for each of 47 French-speaking provinces of Switzerland at about 1888. Our method produces low p-values ($p < 0.0001$) for both the Dip and Silverman tests. Figure 19 in the appendix suggests that this data is indeed clusterable.

The rivers dataset [29] contains the lengths, in miles, of 141 major rivers in North America. Depicted in Figure 26 in the appendix, the dataset exhibits some inherent cluster structure if we allow small clusters. Using our approach, the Silverman test yields $p < 0.0001$. The Dip test results in a $p = 0.2772$, which is above our threshold of 0.05 and indicates a lack of strong cluster structure. Confirming our finding from simulated data, this suggests that the Silverman test may be more appropriate when smaller clusters are of interest, while the Dip test may be desired when the application calls for large clusters and small clusters are better interpreted as outliers.

We now turn to data that appears unclusterable, starting with USArrests [29], depicted in Figure 4. It contains 50 observations and 4 variables: arrests per 100,000 residents for assault, murder, and rape in each of the 50 US states in 1973, as well as percent urban population. Our findings yield $p = 0.9394$ for Dip and $p = 0.1897$ for the Silverman test, both of which are above 0.05, implying that the data is not considered clusterable by our methods.

The next dataset we consider, attitude [17], comes from a survey of the clerical employees of a large financial organization, aggregated from the questionnaires of the approximately 35 employees for each of 30 (randomly selected) departments. The data consists of 30 observations on 7 variables, such as their evaluation of handling of employee complaints and opportunity to learn. The data, visually lacking strong cluster structure as shown in Figure 24 in the appendix, is in fact classified as unclusterable using the Dip test on its pairwise distances, yielding a p-value of 0.9040 for the Dip test and $p = 0.9449$ for the Silverman test.

We now turn to the analysis of higher dimensional unclusterable data, considering the dataset USJudgeRatings [27]. This data set represents lawyers' ratings of state judges in the US Supreme Court, containing 43 observations on 12 numeric variables. As shown in Figure 22 in the appendix, this dataset appears to be unclusterable.

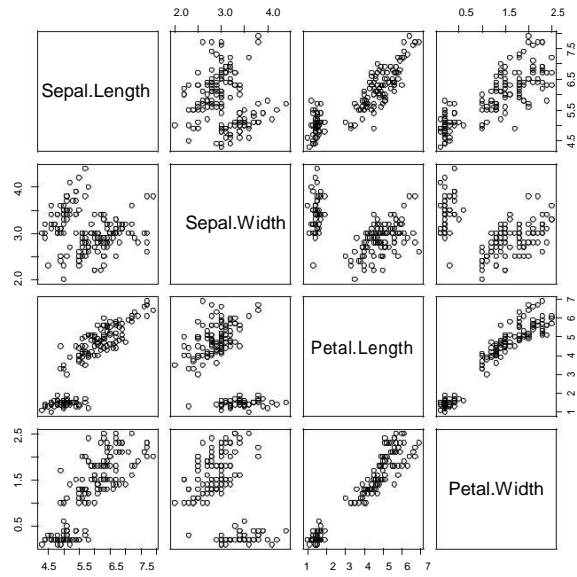

Figure 3: This figure shows 2D projections of the famous iris dataset, which possesses clear cluster structure. Our method for clusterability evaluation attains a p-value of 0 using both the dip and Silverman tests, indicating that iris is clusterable.

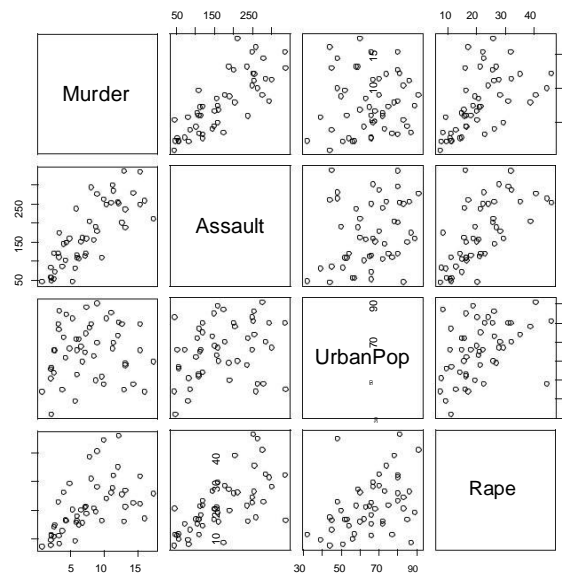

Figure 4: Two-dimensional projections of the USArrests data, suggesting that the data is inherently unclusterable, as confirmed by the p-values of 0.9394 and 0.1897 for the dip and Silverman tests, respectively.

| Dataset | Dip | Silv. |
|---|---|---|
| iris | 0.0000 | 0.0000 |
| swiss | 0.0000 | 0.0000 |
| faithful | 0.0000 | 0.0000 |
| rivers | 0.2772 | 0.0000 |
| trees | 0.3460 | 0.3235 |
| USAJudgeRatings | 0.9938 | 0.7451 |
| USArrests | 0.9394 | 0.1897 |
| attitude | 0.9040 | 0.9449 |
| cars | 0.6604 | 0.9931 |

Table 2: Real data p-values. This table presents the p-values for the dip and Silverman multimodality tests on real datasets from the R Datasets package. Recall that $p < 0.05$ signals clusterable data, which larger values of p signal that data is unclusterable.

Our methods make the same conclusion with $p = 0.9938$ for the Dip test and $p = 0.7451$ for Silverman.

We consider two additional unclusterable datasets in the R datasets package, cars [21] and trees [33]. Both Dip and Silverman correctly classify the data as unclusterable. See Section 4 of the appendix for details.

## 5 Conclusions and discussion

In this paper, we introduce the first practical method for clusterability evaluation that (1) captures how real data is structured, and (2) is computable in low polynomial time. Our approach applies multimodality tests to the (one-dimensional) set of pairwise distances based on original (potentially high-dimensional) data, facilitating easily interpretable results based on sound statistical theory. We demonstrate the utility of our approach on extensive simulations and real data found in the R datasets package.

We emphasize that until now, no experimental study of this type has been done. Previously, notions of clusterability were presented without justification, relying entirely on the formulation of these notions being intuitive. The approach presented in this paper is the first to be supported by empirical results.

Our development of an effective and efficient approach for evaluating clusterability enables several important areas of investigation. Firstly, it allows us to investigate the CDNM thesis, which asks whether "clustering is difficult only when it doesn't matter." This would involve exploring Ben-David's [14] remaining desirable characteristics for notions of clusterability by investigating which algorithms are best suited for uncovering the underlying cluster structure when our approach detects the presence of such structure.

Another avenue for future work is closer investigation of statistical tests of multimodality, such as the application of alternate tests and development of novel multimodality methods that would allow for more fine-grained clusterability analysis. Lastly, we look forward to the application of this new approach for clusterability analysis to practical applications as well as for theoretical investigation, where it will for the first time allow for bridging theory and practice by testing theoretical results on real data.